\documentclass[journal]{IEEEtran}

\ifCLASSINFOpdf
\else
\fi
\usepackage{amsmath}

\usepackage{amssymb,xspace,intmacros,subfigure,amsfonts}

\usepackage[final]{graphicx}

\newcommand{\wb}{{\textbf{w}}}

\newcommand{\ub}{{\textbf{u}}}

\newcommand{\Cb}{{\textbf{C}}}

\newcommand{\Ab}{{\textbf{A}}}

\newcommand{\Ib}{{\textbf{I}}}

\newcommand{\lzero}{$\ell^{0}$}
\newcommand{\lone}{$\ell^{1}$}

\begin{document}
%
\title{Sparse Diffusion Steepest-Descent for One Bit Compressed Sensing in Wireless Sensor Networks}
%
%
%

\author{Hadi~Zayyani,
Mehdi~Korki,~\IEEEmembership{Student Member,~IEEE,}
        and~Farrokh~Marvasti,~\IEEEmembership{Senior Member,~IEEE}
\thanks{H. Zayyani is with the Department
of Electrical and Computer Engineering, Qom University of Technology, Qom, Iran (e-mail: zayyani2009@gmail.com).}
\thanks{M. Korki is with the Department
of Telecommunications, Electrical, Robotics and Biomedical Engineering, Swinburne University of Technology, Hawthorn,
3122 Australia (e-mail: mkorki@swin.edu.au).}
\thanks{F. Marvasti is with the Department
of Electrical Engineering, Sharif University of Technology, Tehran, Iran (e-mail: marvasti@sharif.edu).}

}

%
%

\markboth{IEEE Signal Processing Letters,~Vol.XX, No.X}%
{Shell \MakeLowercase{\textit{et al.}}:}
%



\maketitle

\begin{abstract}
This letter proposes a sparse diffusion steepest-descent algorithm for one bit compressed sensing in wireless sensor networks. The approach exploits the diffusion strategy from distributed learning in the one bit compressed sensing framework. To estimate a common sparse vector cooperatively from only the sign of measurements, steepest-descent is used to minimize the suitable global and local convex cost functions. A diffusion strategy is suggested for distributive learning of the sparse vector. Simulation results show the effectiveness of the proposed distributed algorithm compared to the state-of-the-art non distributive algorithms in the one bit compressed sensing framework.
\end{abstract}


 \ifCLASSOPTIONpeerreview
 \begin{center} \bfseries EDICS: SAS-ADAP \end{center}
 \fi
%
\IEEEpeerreviewmaketitle

\section{Introduction}
%
%
%
%

\IEEEPARstart{W}{e} consider the problem of distributed estimation of a sparse vector in a wireless sensor network from one bit measurements. In a fully distributed wireless sensor network and in the one bit compressed sensing framework, a set of nodes collect the signs of the linear random measurements of a common sparse vector. It is aimed to estimate the sparse vector cooperatively and distributively.

One bit compressed sensing is the extreme case of quantized compressed sensing, where a sparse vector is estimated from only the sign of the linear random measurements \cite{BoufB08}--\nocite{JacqLBB13}\nocite{YanY12}\nocite{PlanV13}\nocite{Li15}\cite{Chen15}. In the compressed sensing (CS) framework \cite{Dono06,CandT06}, a sparse vector $\wb_0$ with a few large coefficients among many zero (or near zero) coefficients, is estimated by only a small number of linear random measurements. Classical CS neglects the quantization process and assumes that the measurements are real continuous valued. However, in practice the measurements should be quantized to some discrete levels. This is known as quantized compressed sensing. In the extreme case, there are only two discrete levels. This is called one bit compressed sensing and it has gained much attention in the research community recently \cite{BoufB08}--\cite{Li15} specially in wireless sensor networks \cite{Chen15}. In the one bit compressed sensing framework, it is proved that an accurate and stable recovery can be achieved by using only the sign of linear measurements \cite{JacqLBB13}. Many algorithms have been designed to solve the problem of one bit compressed sensing such as renormalized fixed-point iteration (RFPI) algorithm \cite{BoufB08}, binary iterative hard thresholding (BIHT) algorithm \cite{JacqLBB13}, adaptive outlier pursuit (AOP) algorithm \cite{YanY12} and One bit Bayesian compressed sensing (BCS) \cite{Li15} to name a few.

In this paper, we focus on the distributive and cooperative estimation of the sparse vector of interest from the signs of the random linear measurements of all the nodes in the wireless sensor network. It is common in the literature to estimate a vector parameter (possibly sparse) from noisy measurements of all the nodes by relying solely on in-network processing \cite{LorS13}. There are numerous algorithms that have been proposed for such problems. The distributed strategy of these algorithms are divided in three main categories: incremental, consensus and diffusion \cite{Sayed14}. In the incremental strategy \cite{Sayed14,LopS07}, each node communicates only with one neighbor node at a time over a cyclic path. Finding a cyclic path which contains all the nodes is an NP-hard problem and the cyclic trajectory is prone to failures \cite{LorS13}. Therefore, other strategies are preferred. In the consensus strategy \cite{Sayed14,Kar09}, at each iteration, every node (or agent) performs two steps: it aggregates the iterates from its neighbors and subsequently updates this aggregate value by the gradient vector evaluated at its existing iterate \cite{Sayed14}. This strategy suffers from the problem of asymmetry which can cause an unstable growth in the state of the network \cite{Sayed14}. Hence, the diffusion strategy \cite{Sayed14,LopS08,CatS10} which removes the asymmetry and not prone to failures, is used. In this strategy, information is processed on the fly by all nodes and the data is diffused across the network using a real-time sharing mechanism \cite{LorS13}.

Motivated by some work that uses distributed estimation in the compressed sensing framework \cite{LopS08,Xu15,ChouT12} and by the distributed diffusion strategy in the networks, we use this strategy for distributed estimation of the sparse vector in the one bit compressed sensing framework. First, two global and local cost functions are defined for the one bit compressed sensing problem. It is proved that if the sparse regularization term is convex, the cost functions are also convex. Hence, a simple steepest-descent method is used for their minimization. Second, two versions of the diffusion strategy (combine then adapt (CTA) and adapt then combine (ATC)) are suggested for the cooperative and distributive estimation of the sparse vector. Experimental results show that the proposed distributed algorithm significantly outperforms the single sensor steepest-descent algorithm. Also, centralized global steepest-descent algorithm outperforms the distributed algorithm. More importantly, the distributed algorithm outperforms the one bit Bayesian compressed sensing (BCS) \cite{Li15} which is the most recent non distributive algorithm in one bit compressed sensing.

\section{Problem formulation}
We consider an ad-hoc wireless sensor network consisting of $N$ nodes (or agents) that are distributed over a region. At every time instant $i$, every node $k$ collects a binary measurement $d_k(i)$ which is the sign of the noisy linear random measurement of a common sparse vector $\wb_o$, i.e.
\begin{equation}
d_k(i)=\mathrm{sign}(\ub_{k,i}\wb_o+v_k(i)),\quad 1\le k\le N,\quad 1\le i\le I
\end{equation}
where $\ub_{k,i}$ is a $1\times M$ random measurement vector and $v_k(i)$ is the measurement Gaussian noise with zero mean and variance $\sigma^2_{v,k}$, independent of $\ub_{l,j}$ for all $l$ and $j$, and independent of $v_l(j)$ for $l\neq k$ and $i\neq j$. The objective of the wireless sensor network is to use the collected data $\{d_k(i),\ub_{k,i}\}$ to estimate the common sparse vector $\wb_o$ in a distributed manner.

\section{Sparse distributed estimation in one bit compressed sensing}
\subsection{Cost Functions}
Similar to \cite{LorS13}, the cooperative and distributive estimation problem in the one bit compressed sensing framework can be regarded as the minimization of the following global cost function:
\begin{equation}
J^{\mathrm{glob}}(\wb)=\sum_{k=1}^N\mathrm{E}(d_k(i)-\mathrm{sign}(\ub_{k,i}\wb))^2+\gamma\mathrm{f}(\wb)
\end{equation}
where $\mathrm{E}$ denotes the expectation operator, and $\mathrm{f}(\wb)$ is a real-valued convex regularization function weighted by the parameter $\gamma>0$, enforcing sparsity of the solution. To avoid complex nonlinear expectation, we replace the expectation with the average time sample. Also, to have a continuous cost function, the sign function is approximated by an S-shaped logistic function $\mathrm{S}(x)=\frac{1-\mathrm{exp}(-x)}{1+\mathrm{exp}(-x)}$. Therefore, the new global cost function is defined as
\begin{equation}
\label{eq: glob}
J^{\mathrm{glob}}(\wb)=\sum_{k=1}^N\sum_{i=1}^I(d_k(i)-\mathrm{S}(\ub_{k,i}\wb))^2+\gamma\mathrm{f}(\wb)
\end{equation}

The minimization of the global cost function in (\ref{eq: glob}) can be solved in a centralized manner. In this method, all the nodes send their data $\{d_k(i),\ub_{k,i}\}$ to a fusion center to collectively process the data. This requires transmitting data between nodes and the fusion center, which demands more power and bandwidth resources. Moreover, centralized approach is prone to fusion center failure. Therefore, the distributed solutions, where each node communicates with its neighbors and signal processing is distributed among all nodes, are usually preferred. In this case, even if some nodes fails, the entire distributed estimation does not collapse. Towards that end, following the approaches outlined in \cite{Sayed14}, a local cost function can be expressed as
\begin{equation}
\label{eq: loc}
J_k^{\mathrm{loc}}(\wb)=\sum_{l\in \mathbb{N}_k}c_{l,k}\sum_{i=1}^I(d_k(i)-\mathrm{S}(\ub_{k,i}\wb))^2+\frac{\gamma}{N}\mathrm{f}(\wb)
\end{equation}
where $c_{l,k}=[\Cb]_{l.k}$ is the weight element of an $N\times N$ matrix $\Cb$ so that
\begin{equation}
c_{l,k}>0 \quad\mathrm{if}\quad l\in\mathbb{N}_k,\quad \sum_{l=1}^Nc_{k,l}=1
\end{equation}
where $\mathbb{N}_k$ is the neighborhood set of node $k$. Each coefficient $c_{l,k}$ represents a weight value that node $k$ assigns to the received information from its neighbor $l$ \cite{Sayed14,LorS13}.

The global cost function is the summation of the local cost functions defined in (\ref{eq: loc})  \cite{Sayed14}:
\begin{equation}
\label{eq: sum}
J^{\mathrm{glob}}(\wb)=\sum_{k=1}^N J_k^{\mathrm{loc}}(\wb)
\end{equation}
Compared to \cite{LorS13}, with the definition in (\ref{eq: loc}), we enforce the sparsity for all the local processors in addition to global processor. Moreover, the common local minimizer of $J_k^{\mathrm{loc}}(\wb)$ is also a local minimizer of $J^{\mathrm{glob}}(\wb)$, due to the definition in (\ref{eq: sum}). It is straightforward to prove that both the global and local cost functions defined in (\ref{eq: glob}) and (\ref{eq: loc}) are convex cost functions assuming the convexity of sparse regularization function $\mathrm{f}(\wb)$. The proof is postponed to the appendix. Because of the convexity, the global and local minimizers of (\ref{eq: glob}) and (\ref{eq: loc}) are the same. Hence, enforcing (\ref{eq: sum}) requires that the common global minimizer of the local cost function is the same as the global minimizer of the global cost function.

\subsection{Sparse Diffusion Steepest-Descent Algorithm}
\label{sec: regfun}
Since the global and local cost functions are convex, global minimizer can be obtained by simple steepest-descent algorithm. The centralized solution via steepest-descent is
\begin{equation}
\wb^{glob}_r=\wb^{glob}_{r-1}-\mu^{glob}\nabla_{\wb}J^{glob}(\wb^{glob}_{r-1}),
\end{equation}
where $r$ is the iteration index and $\nabla_{\wb}J^{glob}(\wb)$ is the gradient vector of $J^{glob}(\wb)$ with respect to $\wb$. The elements of the gradient $[\nabla_{\wb}J^{glob}(\wb)]_j=\frac{\partial}{\partial w_j}J^{glob}(\wb)$ are
\begin{equation}
\sum_{k=1}^N\sum_{i=1}^I\frac{\partial}{\partial w_j}(d_k(i)-\mathrm{S}(\ub_{k,i}\wb))^2+\frac{\gamma}{N}\frac{\partial f(\wb)}{\partial w_j},
\end{equation}
where we have 
\begin{equation}
\frac{\partial}{\partial w_j}(d_k(i)-\mathrm{S}(\ub_{k,i}\wb))^2=-2(d_k(i)-\mathrm{S}(\ub_{k,i}\wb))u_{k,i,j}\mathrm{S}^{'}(\ub_{k,i}\wb)
\end{equation}
    Therefore, the gradient element $[\nabla_{\wb}J^{glob}(\wb)]_j$ is equal to
\begin{equation}
\sum_{k=1}^N\sum_{i=1}^I -2(d_k(i)-\mathrm{S}(\ub_{k,i}\wb))u_{k,i,j}\mathrm{S}^{'}(\ub_{k,i}\wb)+\frac{\gamma}{N}\frac{\partial f(\wb)}{\partial w_j}
\end{equation}

To distributively estimate the sparse vector, a diffusion strategy is suggested which uses the steepest-descent for the adaptation step. Two versions of the diffusion steepest-descent algorithm are adapt then combine (ATC) and combine then adapt (CTA) which can be represented as
\begin{equation}
\mathrm{Diffusion SD-ATC}:\Bigg\{\begin{array}{cc}
                    \wb^{loc}_{r,k}=\wb^{loc}_{r-1,k}-\mu_k\nabla_{\wb}J_k^{loc}(\wb^{loc}_{r-1,k}), \\
                    \wb^{loc}_{r,k}=\sum_{l\in \mathbb{N}_k}a_{l,k} \wb^{loc}_{r,l},
                  \end{array}
\end{equation}
\begin{equation}
\mathrm{Diffusion SD-CTA}:\Bigg\{\begin{array}{cc}
                    \wb^{loc}_{r-1,k}=\sum_{l\in \mathbb{N}_k}a_{l,k}\wb^{loc}_{r-1,l}, \\
                   \wb^{loc}_{r,k}=\wb^{loc}_{r-1,k}-\mu_k\nabla_{\wb}J_k^{loc}(\wb^{loc}_{r-1,k}),
                  \end{array}
\end{equation}
where $a_{l,k}$ is the non-negative combination elements of a combination matrix $\Ab$ which satisfies \cite{Sayed14}:
\begin{equation}
a_{l,k}>0\quad\mathrm{if}\quad l\in\mathbb{N}_k,\quad \sum_{l=1}^Na_{l,k}=1,
\end{equation}
and the gradient element of the local cost function $[\nabla_{\wb}J_k^{loc}(\wb)]_j$ is equal to
\begin{equation}
\sum_{l\in\mathbb{N}_k}c_{l,k}\sum_{i=1}^I -2(d_l(i)-\mathrm{S}(\ub_{l,i}\wb))u_{l,i,j}\mathrm{S}^{'}(\ub_{l,i}\wb)+\frac{\gamma}{N}\frac{\partial f(\wb)}{\partial w_j}.
\end{equation}

The combination coefficients $c_{l,k}$ and $a_{l,k}$ are design parameters determined by the combination policy. Various static combination policies have been suggested such as uniform rule, Laplacian rule and metropolis rule \cite{Sayed14}.

For the sparse regularization function $\mathrm{f}(\wb)$, some functions have been suggested in \cite{LorS13}. One can use \lone-norm $\mathrm{f}_1(\wb)=||\wb||_1=\sum_{m=1}^M|w_m|$ or weighted \lone-norm $||\wb||_0\approx\sum_{m=1}^M\frac{|w_m|}{\eps+|w_m|}$. In addition, we use the smoothed \lzero-norm which is $||\wb||_0\approx\sum_{m=1}^M (1-{\mathrm{e}}^{-\frac{w_m^2}{2\sigma^2}})$ with a small value of $\sigma$ \cite{Mohi09}.

\section{Simulation Results}
In this section, we provide experimental results to illustrate the performance of the diffusion steepest-descent (SD) algorithm. We consider a connected network composed of 10 nodes. The topology of the network is shown in Fig~1. The size of the sparse vector $\wb_o$ is $M=20$. The sparse vector is selected as a Bernoulli-Gaussian (BG) model with activity probability $p=0.2$ which means 20\% of the coefficients are non zero. The variance of the active coefficients is selected as $\sigma^2_w=1$. The number of time samples is selected as $I=40$. The measurement signal $\ub_{k,i}$ is a $1\times 20$ vector with zero mean white Gaussian distributed elements with covariance matrix $\sigma^2_{u,k}\Ib$ and $\sigma_{u,k}=1$. The measurement noise $v_k(i)$ is white Gaussian with $\sigma^2_{v,k}\Ib$ as the covariance matrix with $\sigma_{v,k}=0.01$. For the sparse regularization function, we use \lone-norm with the sparsity parameter $\gamma=10$.

\begin{figure}[tb]
\begin{center}
\includegraphics[width=6cm]{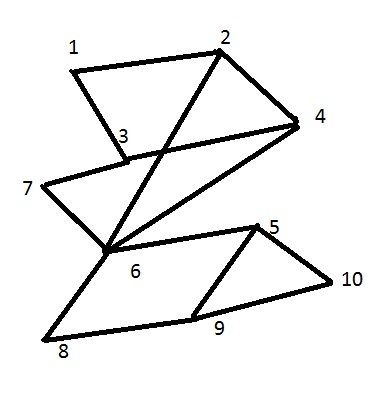}
\caption{Topology of the wireless sensor network.}
\end{center}
\label{fig1}
\end{figure}

The first experiment aims to show the convergence behavior of various algorithms. For performance metric, similar to \cite{LorS13}, we use mean square deviation (MSD) defined as $\mathrm{MSD}(dB)=20\mathrm{log}(||\wb-\wb_o||_2)$. Figure~2 shows the MSD curves versus iteration index for 5 different cooperative algorithms and 2 non cooperative algorithms in the one bit compressed sensing. The cooperative algorithms are centralized steepest-descent (SD), diffusion LMS: ATC \cite{LorS13}, single sensor SD\footnote{An steepest-descent applied only to the local data of sensor 1}, diffusion SD: ATC and diffusion SD: CTA which are proposed in this paper. The two single sensor and non cooperative algorithms are BIHT \cite{JacqLBB13} and one bit BCS \cite{Li15}, whose final MSD performances are also illustrated. In the simulations, we use the same value of $\mu=0.01$ for all step sizes. The results are averaged over 50 independent trials with different sparse vector, measurement vectors and measurement noise. Similar to \cite{LorS13}, we use the matrix $\Cb=\Ib$ which implies that the diffusion algorithms do not exchange the measurements. For the combination matrix $\Ab$, we use the uniform combination policy which simply averages the estimates from the neighboring nodes so that $a_{l,k}=\frac{1}{|\mathbb{N}_k|}$ for all $l$. Figure~2 shows that the best algorithms are the centralized SD and diffusion LMS. Note that the centralized SD uses all the sign data of nodes in a fusion center and the diffusion LMS utilizes the unquantized real valued data. Among the proposed diffusion algorithms, diffusion ATC outperforms diffusion CTA which is consistent with the results reported in \cite{CatS10}. The figure also demonstrates the benefit of cooperation. When the steepest descent is applied to the local data of a single sensor, the final MSD is about -6dB, while the diffusion steepest-descent reaches the final MSD of -20dB, which shows a performance gain of 14dB. It is also seen that the proposed diffusion algorithms outperform the one bit BCS algorithm which is the best non cooperative algorithm in the one bit compressed sensing framework. Moreover, the diffusion LMS applied to the real valued data \cite{LorS13} outperforms the diffusion SD applied to the binary data. Diffusion LMS \cite{LorS13} converges faster than the proposed diffusion SD while it also has a slightly lower final MSD (2dB). This is because diffusion LMS exchanges the real valued data, while diffusion SD exchanges the binary data, which results in higher complexity of the nodes of the wireless sensor network.

\begin{figure}[tb]
\begin{center}
\includegraphics[width=10cm]{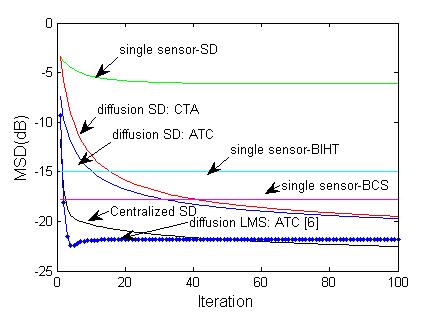}
\caption{MSD of various algorithms. Algorithms are centralized steepest-descent (SD), diffusion LMS: ATC \cite{LorS13}, single-sensor SD, diffusion SD: ATC, diffusion SD: CTA, BIHT and one bit BCS.}
\end{center}
\label{fig2}
\end{figure}

The second experiment investigates the effect of the regularization function in the proposed diffusion ATC algorithm. The parameters are the same as the first experiment. Figure~3 shows the MSD curves versus iteration index for three different regularization function which are \lone-norm, weighted \lone-norm (with $\eps=1e^{-10}$) and smoothed \lzero-norm (with $\sigma=1e^{-3}$), which are introduced in Section \ref{sec: regfun}. There we observe that the best regularization function is the \lone-norm.

\begin{figure}[tb]
\begin{center}
\includegraphics[width=10cm]{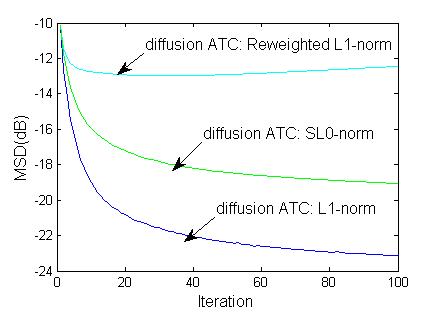}
\caption{MSD of diffusion ATC algorithm with various sparse regularization functions. The sparse regularization functions are \lone-norm, reweighted \lone-norm and smoothed \lzero-norm (SL0).}
\end{center}
\label{fig3}
\end{figure}

\section{Conclusion}
In this letter, we have proposed a family of diffusion steepest-descent algorithms for distributed estimation of a sparse vector from the sign of linear measurements in the one bit compressed sensing framework. The convex global and local cost functions are properly defined for the problem. Then, the steepest-descent algorithm is used to obtain the common global minimizer. Two diffusion strategies are also suggested for distributed estimation in the wireless sensor network. Simulation results show the effectiveness of the algorithms to estimate the sparse vector and the superiority of the proposed diffusion algorithms over the single sensor and one bit BCS which is the best non cooperative algorithm.

\appendix  


To verify the convexity of the global and local cost functions, because of the similarity it suffices to prove the convexity of the global cost function. Assuming the convexity of the sparse regularization function, we should prove the convexity of $T(\wb)=\sum_{k=1}^N\sum_{i=1}^I(d_k(i)-\mathrm{S}(\ub_{k,i}\wb))^2$. It can be shown that the second partial derivative $\frac{\partial^2T(\wb)}{\partial w^2_j}$ is
\begin{equation}
\sum_{k=1}^N\sum_{i=1}^I2u^2_{k,i,j}[-(d_k(i)-\mathrm{S}(\ub_{k,i}\wb))\mathrm{S}^{''}(\ub_{k,i}\wb)+{\mathrm{S}^{'}}^2(\ub_{k,i}\wb)].
\end{equation}

If $x=\ub_{k,i}\wb>0$ then $d_k(i)=1$ and it can be shown that $[-(d_k(i)-\mathrm{S}(\ub_{k,i}\wb))\mathrm{S}^{''}(\ub_{k,i}\wb)+{\mathrm{S}^{'}}^2(\ub_{k,i}\wb)]=\frac{4\mathrm{e}^{-2x}(2-\mathrm{e}^{-x})}{(1+\mathrm{e}^{-x})^4}>0$, and hence $\frac{\partial^2T(\wb)}{\partial w^2_j}>0$. On the other hand, when $x=\ub_{k,i}\wb<0$ we have $d_k(i)=-1$. Then we have $[-(d_k(i)-\mathrm{S}(\ub_{k,i}\wb))\mathrm{S}^{''}(\ub_{k,i}\wb)+{\mathrm{S}^{'}}^2(\ub_{k,i}\wb)]=\frac{4\mathrm{e}^{-x}(2\mathrm{e}^{-x}-1)}{(1+\mathrm{e}^{-x})^4}>0$. Therefore, we have $\frac{\partial^2T(\wb)}{\partial w^2_j}>0$. By proving that the second derivative is always positive, the proof of convexity is completed.



\ifCLASSOPTIONcaptionsoff
  \newpage
\fi

\end{document}